\documentclass[runningheads]{llncs}
\usepackage[T1]{fontenc}
\usepackage{graphicx}
\usepackage{booktabs}
\usepackage{subcaption}
\begin{document}
\title{How Effective are Self-Supervised Models for Contact Identification in Videos}
\titlerunning{How Effective are Self-Supervised Models}

\author{Malitha Gunawardhana*\inst{1} \and
Limalka Sadith\inst{1,2} \and
Liel David\inst{3}\and
Daniel Harari\inst{3} \and
Muhammad Haris Khan\inst{1}
}
\authorrunning{Gunawardhana et al.}
\institute{Mohamed bin Zayed University of Artificial Intelligence: MBZUAI, UAE \and
University of Moratuwa, Sri Lanka \and
Weizmann Institute of Science, Israel\\ {*}Corresponding author. Email: \email{malithagunawardhana96@gmail.com}
}

\maketitle              %
\begin{abstract}
The exploration of video content via Self-Supervised Learning (SSL) models has unveiled a dynamic field of study, emphasizing both the complex challenges and unique opportunities inherent in this area. Despite the growing body of research, the ability of SSL models to detect physical contacts in videos remains largely unexplored, particularly the effectiveness of methods such as downstream supervision with linear probing or full fine-tuning. This work aims to bridge this gap by employing eight different convolutional neural networks (CNNs) based video SSL models to identify instances of physical contact within video sequences specifically. The Something-Something v2 (SSv2) and Epic-Kitchen (EK-100) datasets were chosen for evaluating these approaches due to the promising results on UCF101 and HMDB51, coupled with their limited prior assessment on SSv2 and EK-100. Additionally, these datasets feature diverse environments and scenarios, essential for testing the robustness and accuracy of video-based models. This approach not only examines the effectiveness of each model in recognizing physical contacts but also explores the performance in the action recognition downstream task. By doing so, valuable insights into the adaptability of SSL models in interpreting complex, dynamic visual information are contributed.

\end{abstract}

\section{Introduction}

Automated video analysis has significantly evolved from basic frame-by-frame methods to sophisticated systems that understand the temporal dynamics in videos \cite{donahue2015long,liu2018t,zhou2022survey}. This shift marks a move from analyzing static images to interpreting complex, dynamic scenes, enhancing our understanding of visual content  \cite{wang2019self,jing2018self}. Initially, video analysis relied on manually extracted features and simple models to understand temporal relationships, which worked well in controlled settings but struggled with real-world video complexity \cite{schuldt2004recognizing,wang2013action}. The introduction of deep learning, particularly Convolutional Neural Networks (CNNs), revolutionized video analysis, improving its applicability and performance. However, these advancements depend heavily on large, annotated datasets, which are often costly and scarce, posing scalability and adaptability challenges \cite{hendrycks2019benchmarking,hendrycks2021many}.

The rise of self-supervised learning (SSL) offers a solution by using the data's inherent structure to create learning signals, eliminating the need for extensive annotations. This paradigm is particularly promising in video analysis as it allows models to learn from the data itself, for instance by leveraging the continuity and predictability of visual elements to build robust, context-aware systems \cite{jing2020self}. This research direction promises to learn comprehensive, context-sensitive representations by exploiting temporal coherence \cite{chen2022frame,knights2021temporally}, spatial continuity \cite{wang2023masked} and the predictability of motion patterns \cite{sun2023masked},

Distinguishing itself from image-based SSL methods, video-based SSL confronts unique challenges due to the intrinsic properties of videos \cite{tran2018closer,fernando2017self}. Initially, researchers explored simple ordering-based methods to achieve promising results \cite{noroozi2016unsupervised,wei2019iterative}. Subsequently, various strategies utilizing the Vision Transformer (ViT) \cite{dosovitskiy2020image} frameworks were proposed, introducing new areas to the field. The typical training path for SSL models involves a two-step process: a pretraining stage followed by a fine-tuning stage. During pretraining, models use supervisory signals to capture the intrinsic nature of the data. For instance, in video data, this could involve understanding common motion patterns, recognizing typical backgrounds, and identifying the relationship between different frames.  The trained models are then fine-tuned for specific downstream task such as action recognition and video retrieval. Most models undergo pretraining using the Kinetics-400 dataset \cite{kay2017kinetics}  and are evaluated on the UCF-101 dataset \cite{soomro2012ucf101}, with evaluations focusing on action recognition or video retrieval tasks. Despite achieving notable performance, questions about the generalizability of these methods remain \cite{thoker2022severe}.

The ability to understand video content at intricate level has profound implications across various applications. This nuanced comprehension can substantially enhance surveillance systems by incorporating real-time alerting features that detect specific actions, such as distinguishing between a friendly wave and a distressed hand signal. Furthermore, it can redefine human-computer interaction in the sphere of robotics, enabling systems to interpret and respond accurately to subtle human behaviors. Among the  various challenges in this domain, the accurate identification of physical contact between objects or entities within video frames stands out as particularly critical. This aspect is essential for applications such as physical interaction detection \cite{narasimhaswamy2020detecting}, and the creation of immersive augmented reality experiences \cite{kyriakou2019can}. This paper's focus is to thoroughly evaluate the efficacy of current CNN-based SSL models in identifying contact within videos. Such an examination is pivotal in advancing our understanding of visual content, setting the stage for significant progress in video analysis methodologies without relying heavily on extensively labelled datasets. The structure of this paper is outlined as follows.  First, we offer a comprehensive explanation of the models chosen for evaluation. Next, we describe the experimental setup. We then present the results and discuss their implications. The paper concludes with a summary of the findings in the conclusion section.

\section{Evaluated models}

\begin{table}[b]
\caption{Selected models for evaluation, grouped by type: pretext task-based, contrastive learning-based, generative learning-based, and multi-modal-based.}
\label{tab:summary}
\centering
\begin{tabular}{@{}llr@{}}
\toprule
Model            & Method              & Year \\ \midrule
CTP              & Pretext task        & 2021 \\
RSPNet           & Pretext task            & 2021 \\
TCLR             & Contrastive learning        & 2022 \\
MoCo             & Contrastive learning        & 2021 \\
Pre-Con & Contrastive learning + Pretext task & 2020 \\
VideoMoCo        & Generative learning        & 2021 \\
AVID-CMA         & Multi-modal         & 2021 \\
GDT              & Multi-modal         & 2021 \\

\bottomrule
\end{tabular}

\end{table}

For an unbiased and objective assessment, our evaluation exclusively encompasses CNN-based models only. These models have all been pre-trained on the Kinetic-400 dataset, ensuring a consistent foundation across the board. To further enhance the fairness and comparability of our evaluation, each model employs the R(2+1)D-18   \cite{tran2018closer} architecture as its backbone. This specific architecture choice aligns with our goal to maintain a uniform structure across models, minimizing variability that could arise from differing network designs. We have selected a total of eight CNN-based models that utilize SSL techniques for this evaluation.

Selected models are AVID-CMA \cite{morgado2021audio}, 
Catch the Patch (CTP) \cite{wang2021unsupervised}, 
GDT \cite{patrick2020multi}, 
MoCo \cite{chen2020improved},VideoMoCo \cite{pan2021videomoco} , 
Pretext-Contrast (Pre-Con) \cite{tao2020pretext},
RSPNet \cite{chen2021rspnet}, and
TCLR \cite{dave2022tclr}.

AVID-CMA \cite{morgado2021audio} is a multi-modal learning framework that takes advantage of both video and audio data. This method employs Audio-Visual Instance Discrimination (AVID) to foster a cross-modal similarity metric, effectively pairing video and audio instances that coexist. Furthermore, Cross-Modal Agreement (CMA) enhances this approach by clustering videos that exhibit strong similarity across both video and audio dimensions, thereby refining the selection of positive and negative samples for training.

Inspired by how the human eyes work during childhood, CTP \cite{wang2021unsupervised} proposed an SSL pretext task known as "Catch the Patch".  The method involves tracking patches across video frames to learn consistent and robust feature representations. The core idea is that the temporal consistency of appearances and motions in videos offers a rich, unsupervised signal for learning. By focusing on patches rather than whole frames or objects, the method captures fine-grained details.

GDT \cite{patrick2020multi} is also a multi-modal learning with contrastive learning. Authors generalize contrastive learning to a wider set of transformations and their compositions, aiming for invariance or distinctiveness in image representations. Previous work has shown that the choice and composition of transformations are crucial for performance in contrastive learning. However, these choices have been mostly driven by intuition, lacking formal understanding and generalization.  The authors propose a formal analysis of composable transformations in contrastive learning, providing principles for constructing training batches. They introduce a practical construction that satisfies the requirements of contrastive formulations.

Momentum Contrast or MoCo \cite{chen2020improved} is one of the famous contrastive learning methods that focus on images. The proposed method, which consists of a dynamic dictionary and moving-average encoder, allows for the on-the-fly construction of a large, consistent dictionary, enhancing the effectiveness of contrastive unsupervised learning. VideoMoCo \cite{pan2021videomoco} extends MoCo concepts to videos. 

Rather than utilizing a single pretext task or contrastive learning method, the Pre-Con \cite{tao2020pretext} method combines these two aspects. This joint optimization framework can improve performance rather than using one method.  They use 3D RotNet \cite{jing2018self}, VCP \cite{luo2020video} and VCOP \cite{xu2019self} as the pretext task.

RSPNet \cite{chen2021rspnet} is another pretext task-based method which focuses on using speed for supervision. They use relative speed between two clips rather than using the exact speed.  To ensure the learning of appearance features, RSPNet also introduces an appearance-focused task, where the model is enforced to perceive the appearance difference between two video clips. 

TCLR \cite{dave2022tclr} is another contrastive learning approach designed to emphasize the distinct of features over time. Unlike previous contrastive learning methods for video data, which did not specifically focus on temporal feature distinctiveness, TCLR uses clips from the same video as negative examples to promote diversity across time. The method introduces two innovative loss functions: the local-local temporal contrastive loss, targeting discrimination among non-overlapping clips from the same video, and the global-local temporal contrastive loss, which aims to enhance the temporal variance of features by distinguishing between different time steps within the feature map of a clip.

A summary of these models and their modality (based on \cite{schiappa2023self}) is shown in Table \ref{tab:summary}.

\begin{figure}[h!]
    \centering
    \includegraphics[width=1\linewidth]{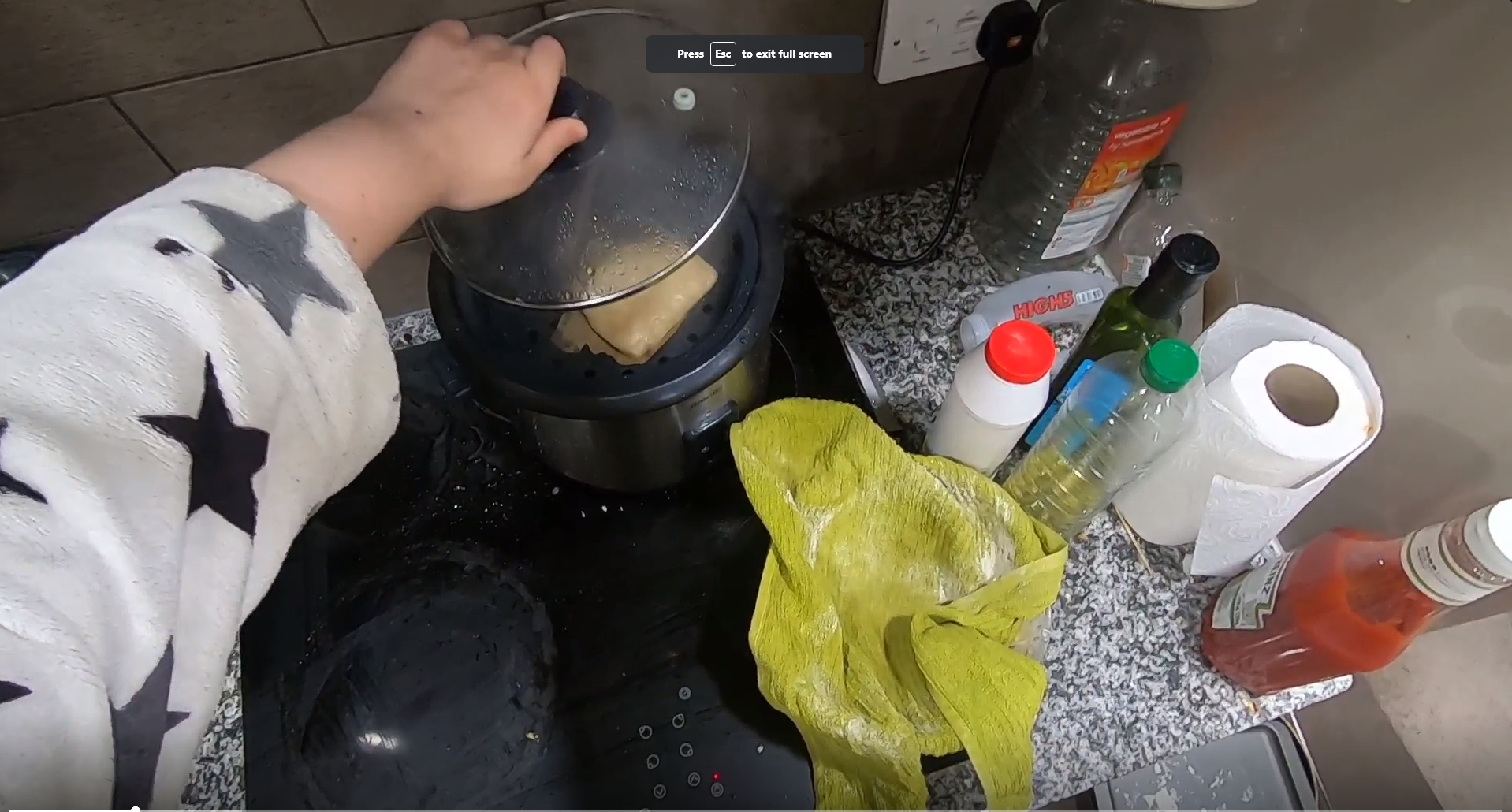}
    \caption{lift lid off rice cooker. verb:- lift off, noun:- lid}
    \label{fig:ek1}
\end{figure}

\begin{figure}[h!]
    \centering
    \includegraphics[width=1\linewidth]{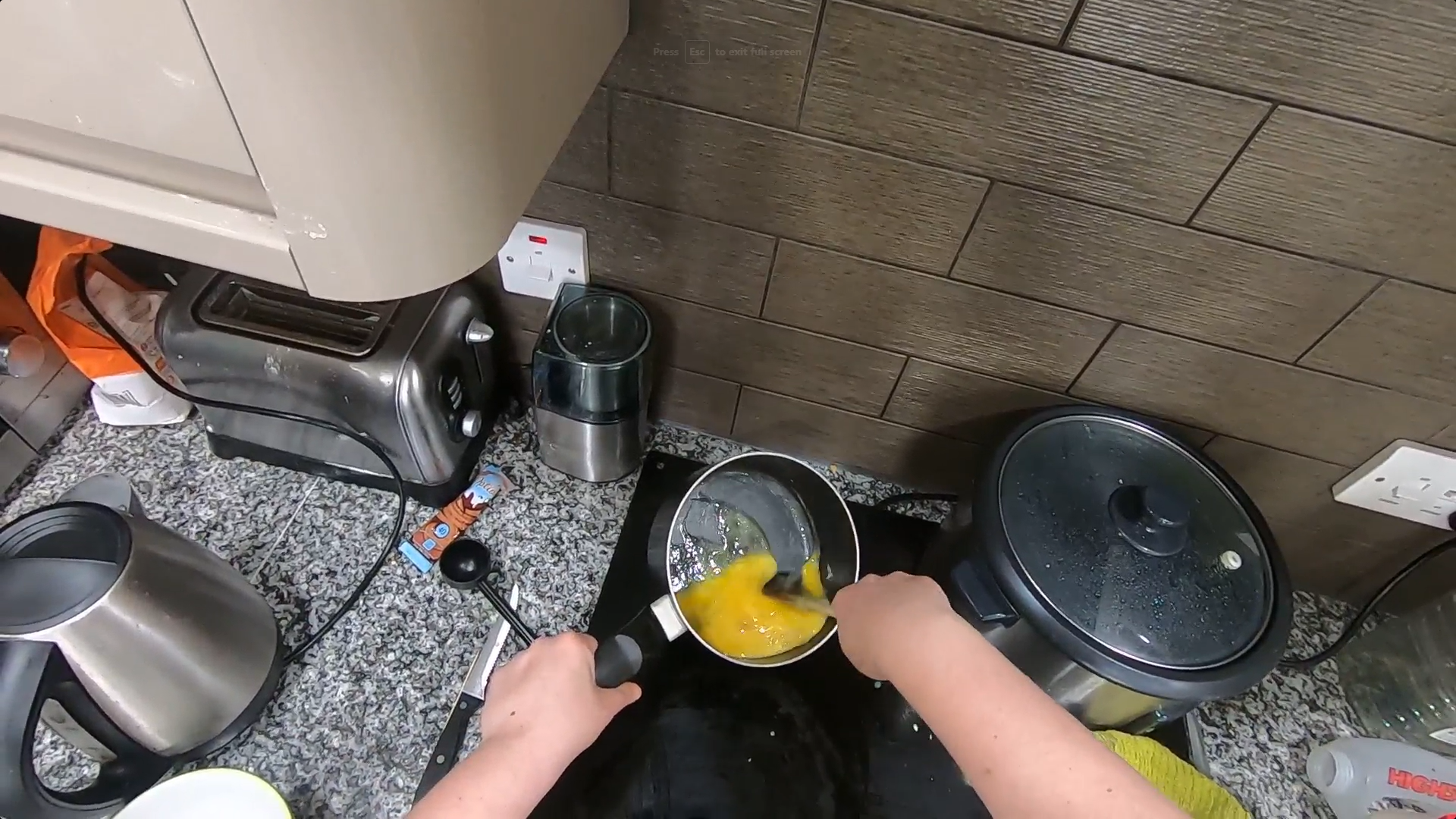}
    \caption{stir egg in pan using spatula, verb:- stir -in, noun:- egg}
    \label{fig:ek2}
\end{figure}

\section{Experimental Setup}
\subsection{Dataset}

We utilize two main datasets. The first dataset is the Something-Something V2 (SSv2) \cite{goyal2017something} dataset. The SSv2 contains 168913 training videos and 24777 validation videos. Each video is assigned to motion-centric action classes known as templates. There are 174 templates, including descriptions like "Holding something next to something" and "Digging something out of something," among others. 

The second dataset is the Epic-Kitchen-100 (EK-100) dataset \cite{Damen2022RESCALING}. This is a large-scale video dataset based on day-to-day activities in the kitchen. The dataset is divided into three main categories based on the annotations. Those are 1) Noun, 2) Verb and 3) Action. Videos are ego-centric and collected from 45 kitchens in four cities. A total of 97 verb classes and 300 noun classes are available in this dataset. Figure \ref{fig:ek1} and Figure \ref{fig:ek2} show an example of the EK-100 video.

These datasets were selected due to their unique characteristics, which align well with the objectives of this study. SSv2 captures a wide range of action categories and temporal dynamics, making it ideal for training models to recognize various contact-based activities. EK100, with its realistic context and ego-centric perspective, captures everyday activities involving numerous object interactions. The rich annotations, including action verbs and nouns, along with temporal segmentation, enhance the dataset's utility in identifying specific contact events.

\begin{figure}[b]
    \centering
    \includegraphics[width=1\linewidth]{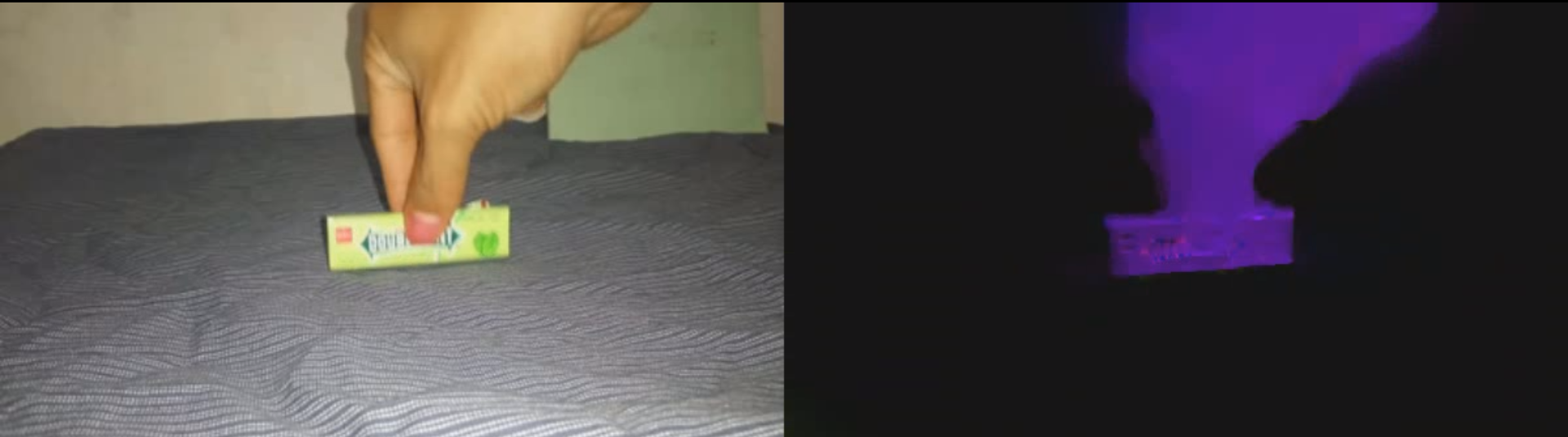}
    \caption{SSv2 True category example}
    \label{fig:ssv21}
\end{figure}

\begin{figure}[ht]
    \centering
    \includegraphics[width=1\linewidth]{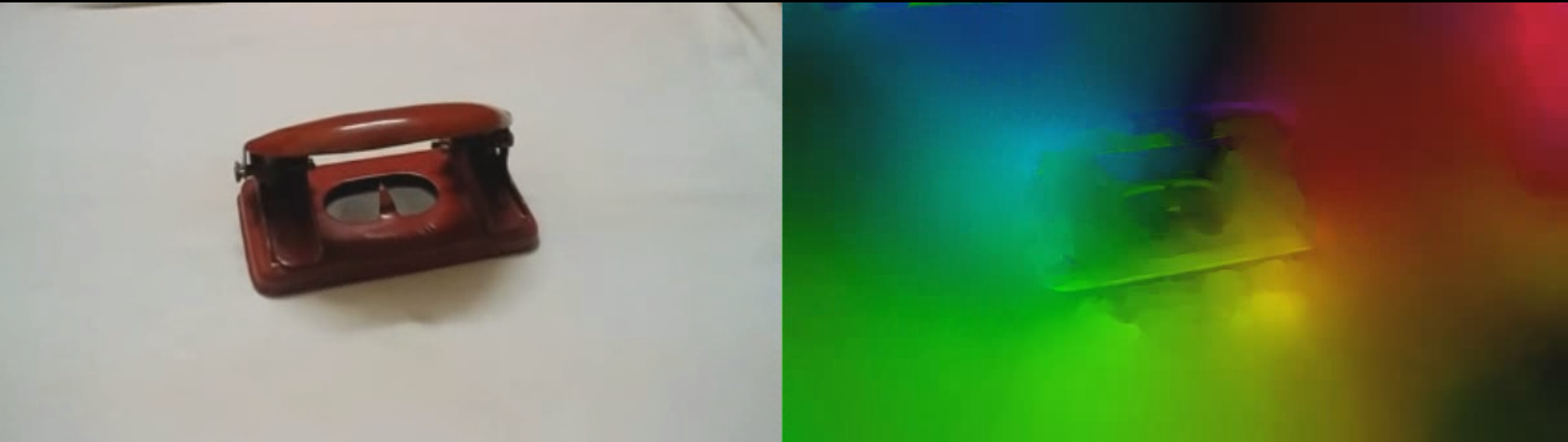}
    \caption{SSv2 False category example}
    \label{fig:ssv22}
\end{figure}

We conduct an exhaustive review and analysis of each video template within SSv2 and each action within EK-100. For the purposes of this analysis, both video templates from SSv2 and actions from EK-100 are collectively referred to as "templates." These templates and videos are categorized into two primary groups: 1) videos depicting human interaction with objects, designated as the "True" category, and 2) videos lacking any human-object interaction, designated as the "False" category. It's important to note that the presence of contact between humans and objects is not immediately apparent through visual inspection alone. As a result, we employed optical flow analysis \cite{perez2013robust} for each videos to discern instances of contact. However, it was observed that within some videos, the presence of contact was inconsistent across videos. Therefore, these videos and templates were excluded from our analysis. Examples of these cases for the SSv2 dataset are shown in Figure \ref{fig:ssv21} and Figure \ref{fig:ssv22}.

\subsection{Implementation Details}

We implement these methods using the PyTorch 1.6.0 framework and train all models on eight Tesla A100 GPUs. Our evaluation methodology draws inspiration from \cite{thoker2022severe}, adopting the same hyperparameters they utilized. For fine-tuning the SSv2 dataset, we employ a batch size of 32 and a learning rate of 0.0001, training each model for a total of 45 epochs. Similarly, for the EK-100 dataset, we maintain the same batch size but adjust the learning rate to 0.0025 and train over 30 epochs. In the context of linear evaluation, the SSv2 dataset, we set a batch size of 64 and a learning rate of 0.01 for 40 epochs, whereas, for EK-100, we reduce the batch size to 32 while retaining a learning rate of 0.0025 over 30 epochs.

\subsection{Evaluation Method}

\begin{figure*}[h!]
\centering
\begin{subfigure}{.5\textwidth}
  \centering
  \includegraphics[width=.9\linewidth]{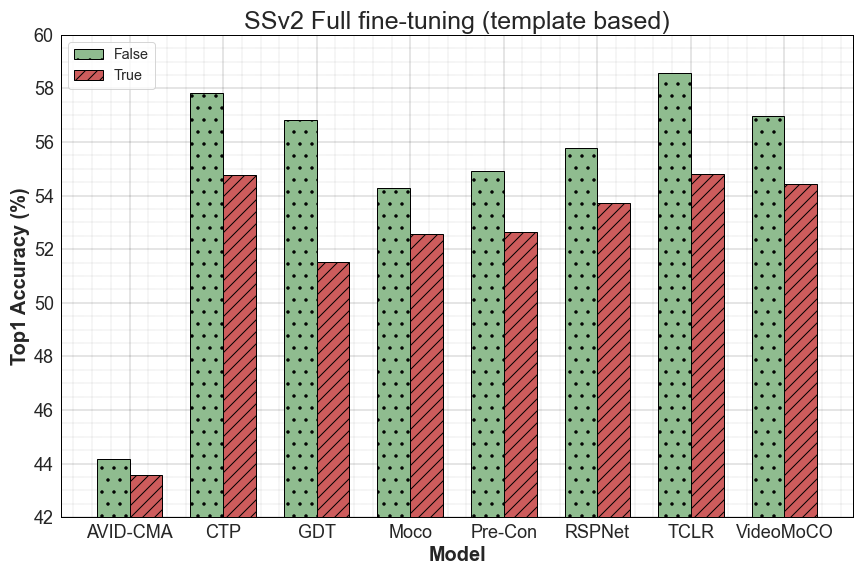}
  \caption{Full fine-tuning}
  \label{fig:ssv2fft}
\end{subfigure}%
\begin{subfigure}{.5\textwidth}
  \centering
  \includegraphics[width=.9\linewidth]{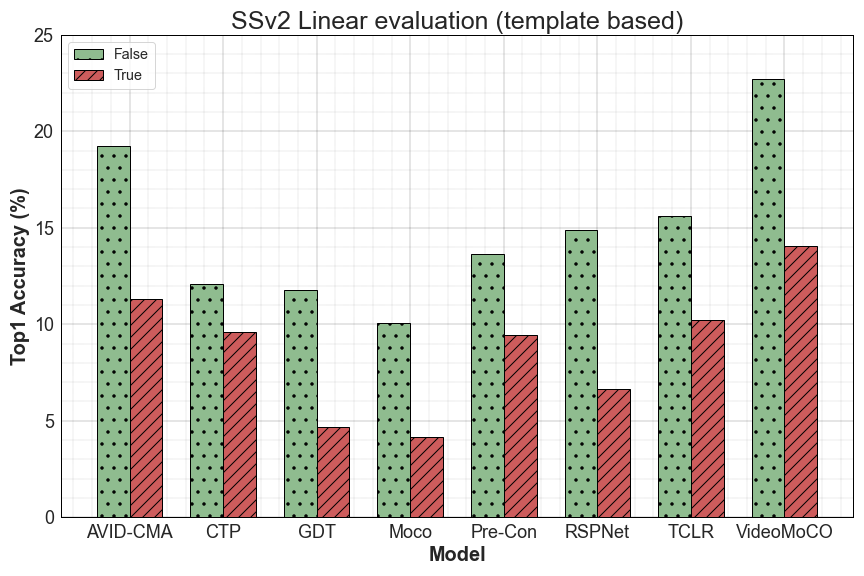}
  \caption{Linear evaluation}
  \label{fig:ssv2linear}
\end{subfigure}

\caption{Template-based evaluation for SSv2 dataset}

\label{fig:ssv2-templatae}
\end{figure*}

\begin{figure*}[h!]
\centering
\begin{subfigure}{.5\textwidth}
  \centering
  \includegraphics[width=.9\linewidth]{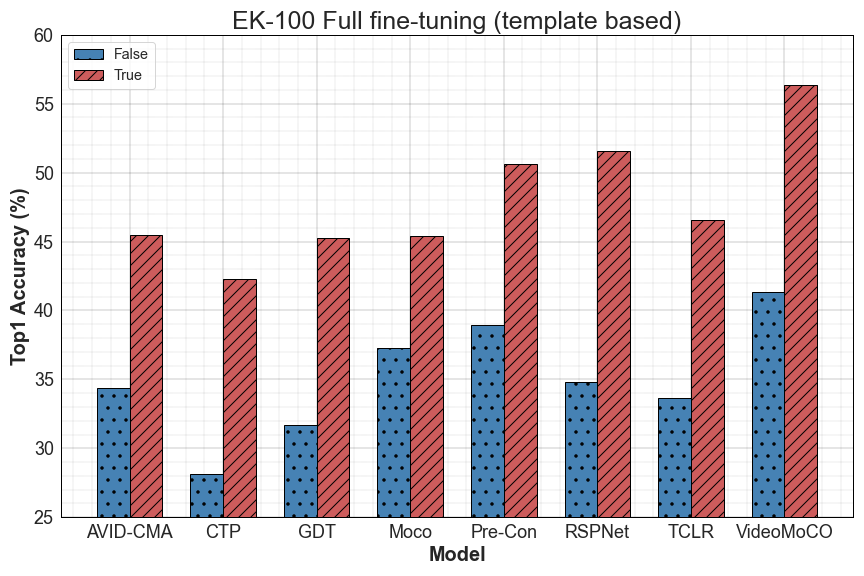}
  \caption{Full fine-tuning}
  \label{fig:ekfft}
\end{subfigure}%
\begin{subfigure}{.5\textwidth}
  \centering
  \includegraphics[width=.9\linewidth]{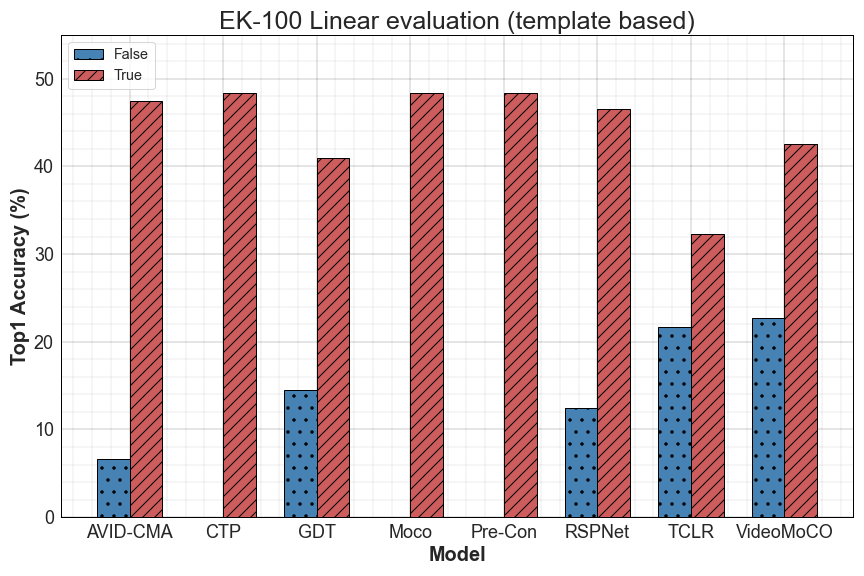}
  \caption{Linear evaluation}
  \label{fig:eklinear}
\end{subfigure}

\caption{Template-based evaluation for EK-100 dataset}

\label{fig:ek-templatae}
\end{figure*}

We assess our models under two main conditions: Overall performance and Template-based performance. Overall performance evaluates the models based on action recognition accuracy across the entire dataset. Template-based performance evaluation focuses on scenarios where there is contact between humans and objects. For overall performance on the SSv2 dataset, we utilize four accuracy metrics: Top-1, Top-5, Mean Top-1, and Mean Top-5. In contrast, the performance of the EK-100 dataset is evaluated using only Top-1 and Mean Top-1 accuracy metrics. During the Template-Based Performance evaluation, we apply only the Top-1 accuracy metric for both datasets. Moreover, only the verb class is considered for Template-based performance evaluation.

Top-1 accuracy can be defined as the correct number of predictions divided by total predictions. Top-5 accuracy is determined by the condition that at least one of the top five predictions with the highest probabilities corresponds to the actual outcome. The mean Top-1 accuracy refers to the average accuracy across all classes or datasets where, for each prediction, only the most probable outcome is considered correct. It's the mean of the Top-1 accuracies for each class or dataset. On the contrary, the mean Top-5 accuracy follows a similar concept. Still, it extends the criteria for a correct prediction to any of the top five most probable outcomes predicted by the model. It is the average of the Top-5 accuracies for each class or dataset.

In addition to the previously mentioned evaluation criteria, we evaluate both datasets under two main conditions. 1) full fine-tuning and 2) linear evaluation. During the full fine-tuning phase, the whole weights are modified throughout the training process. It allows us to modify the pre-trained model with small, incremental changes that are designed to enhance performance on the specific task. Linear evaluation keeps the pre-trained model static and learns a linear layer on top of it to map the model's output to the target task's output. In other words, we freeze the entire network except for the last linear/classification layer.

In the SSv2 dataset, there are a total of 24,777 videos distributed across 97 templates. Of these, 12,620 videos are labeled as 'false' and 7,812 as 'true'. The EK-100 dataset contains 9,668 videos, with 4,001 labeled as 'true' and 3,389 as 'false'. The remaining 4,345 videos in SSv2 and 2278 videos in EK-100 contain both true and false videos, and were excluded from the analysis. A summary of these statistics is presented in Table \ref{tab:videos_templates} for both SSv2 and EK-100 datasets.

\begin{table}
\caption{Number of  Videos for SSv2 and EK-100 datasets}
\centering
\begin{tabular}{ccc}
\toprule
\textbf{Category} & \textbf{SSv2} & \textbf{EK-100} \\ \midrule

True              & 7,812                     & 4001                           \\
False             & 12,620                    &    3389                        \\
Both              & 4,345                     & 2278                           \\ 
Total                & 24,777                    & 9668                          \\
\bottomrule
\end{tabular}
\label{tab:videos_templates}
\end{table}

\section{Results}

Table \ref{tab:ssv2-ff}, Table \ref{tab:ssv2-linear}, Table \ref{tab:EK-100-fft}, and Table \ref{tab:ek-linear} present the results for SSv2 full fine-tuning, SSv2 linear evaluation, EK-100 full fine-tuning, and EK-100 linear evaluation, respectively under overall performance evaluation.

Figure \ref{fig:ssv2-templatae} illustrates the template-based performance on the SSv2 dataset, detailing both full fine-tuning (see Figure \ref{fig:ssv2fft}) and linear evaluation scenarios (refer to Figure \ref{fig:ssv2linear}). Similarly, Figure \ref{fig:ek-templatae} demonstrates the template-based performance for the EK-100 dataset, with a specific focus on full fine-tuning (as shown in Figure \ref{fig:ekfft}) and linear evaluation settings (outlined in Figure \ref{fig:eklinear}). All values are presented as percentage (\%) values in all figures and tables.

\begin{table}
\caption{Overall performance comparison of action recognition accuracy in SSv2 dataset for full fine-tuning evaluation}
\label{tab:ssv2-ff}
\centering
\setlength{\tabcolsep}{5pt} %
\begin{tabular}{@{}llclc@{}}
\toprule
Model & Top-1 & Mean Top-1 & Top-5 & Mean Top-5 \\ \midrule
AVID-CMA & 45.26 & 38.11 & 76.45 & 70.41 \\
CTP & 57.09 & 52.02 & \textbf{84.39} & \textbf{81.15} \\
GDT & 55.15 & 50.50 & 82.18 & 78.72 \\
MoCo & 54.05 & 48.56 & 82.47 & 78.92 \\ 
Pre-Con & 54.63 & 48.91 & 82.46 & 78.82 \\
RSPNet & 55.28 & 50.18 & 82.96 & 79.30 \\
TCLR & \textbf{57.43} & \textbf{52.55} & 83.92 & 80.69 \\
VideoMoco & 56.42 & 51.39 & 83.26 & 80.10 \\
\bottomrule
\end{tabular}

\end{table}

\begin{table}[ht]
\caption{Overall performance comparison of action recognition accuracy in SSv2 dataset for linear evaluation.}
\label{tab:ssv2-linear}
\centering
\setlength{\tabcolsep}{5pt} %
\begin{tabular}{llclc}
\hline
Model                & Top-1  & Mean Top-1 & Top-5  & Mean Top-5 \\
\hline
AVID-CMA             & 16.08 & 12.46     & 38.31 & 31.20\\
CTP                  & 11.15 & 8.33      & 27.55 & 20.23     \\
GDT                  & 8.94  & 7.33      & 25.58 & 21.47     \\
MoCo                 & 7.21  & 5.03      & 22.04 & 16.10\\
Pre-Con     & 11.62 & 8.11      & 30.11 & 22.91     \\
RSPNet               & 11.12 & 8.38      & 30.34 & 24.18     \\
TCLR                 & 13.41 & 9.43      & 33.96 & 25.75     \\
VideoMoCo            & \textbf{19.66} & \textbf{15.46}     & \textbf{43.02 }& \textbf{35.76}     \\
\hline
\end{tabular}

\end{table}

\begin{table}[ht]
\caption{Overall performance comparison of verb, noun, action recognition accuracy in EK-100 dataset for full fine-tuning evaluation.}
\label{tab:EK-100-fft}
\centering
\setlength{\tabcolsep}{3pt} %
\begin{tabular}{@{}lllllll@{}} \toprule
 Model& \multicolumn{2}{c}{Verb}& \multicolumn{2}{c}{Noun}& \multicolumn{2}{c}{Action}\\
\cline{2-7}
& Top-1& Top-5&  Top-1& Top-5&  Top-1& Top-5\\ \midrule
AVID-CMA         & 34.37     & 73.48     & 10.10      & 27.99     & 4.15        & 24.49      \\
CtP              & 30.06     & 71.17     & 8.95      & 24.80      & 3.80& 20.95      \\
GDT              & 34.56     & 75.25     & 12.39     & 30.93     & 6.66        & 27.16      \\
MoCo             & 37.76     & 76.90     & 15.42     & 34.70      & 9.35        & 30.17      \\
Pre-Con & 39.40     & 76.37     & 13.99     & 32.91     & 8.21        & 29.14      \\
RSPNet           & 40.27     & \textbf{78.09 }    & \textbf{18.20}& \textbf{39.55}     & \textbf{11.23}   & \textbf{34.76 }     \\
TCLR             & 34.21     & 75.00& 10.98     & 29.59     & 5.08        & 25.55      \\
VideoMoCo        & \textbf{44.53}     & 77.82     & 16.36     & 36.23     & 10.88       & 31.98      \\
\bottomrule
\end{tabular}
\end{table}

\begin{table}[h!]
\caption{Overall performance comparison of verb, noun, action recognition accuracy in EK-100 dataset for linear evaluation}
\label{tab:ek-linear}
\centering
\setlength{\tabcolsep}{3pt} %
\begin{tabular}{@{}lcccccc@{}} \toprule
 Model& \multicolumn{2}{c}{Verb}& \multicolumn{2}{c}{Noun}& \multicolumn{2}{c}{Action}\\
\cline{2-7}
& Top-1& Top-5&  Top-1& Top-5&  Top-1& Top-5\\ \midrule
AVID-CMA & 21.95 & 66.01 & 3.93 & 19.82 & 0.02 & 14.36 \\
CtP & 20.04 & 62.07 & 4.49 & 17.93 & 0.97 & 12.51 \\
GDT & 22.07 & 64.65 & 5.38 & 18.92 & 0.86 & 14.13 \\
MoCo & 20.04 & 63.21 & 3.93 & 18.89 & 0.01 & 13.3 \\
Pre-Con & 20.04 & 66.03 & 3.93 & 18.59 & 0.01 & 13.26 \\
RSPNet & 23.62 & 67.86 & 6.40& 20.43 & 0.61 & 15.43 \\
TCLR & 23.27 & 69.02 & \textbf{7.16} & \textbf{22.67} &\textbf{1.69} & \textbf{17.41} \\
VideoMoCo & \textbf{25.57} & \textbf{70.06} & 6.00& 20.48 & 0.80& 16.18 \\
\hline
\end{tabular}

\end{table}

When evaluating overall performance across both datasets under two assessment conditions (full fine-tuning and linear evaluation), it's apparent that no single method consistently outperforms the others. Specifically, within the SSv2 dataset's full fine-tuning category (Table \ref{tab:ssv2-ff}), TCLR achieves the highest Top-1 accuracy, whereas CTP secures the best Top-5 accuracy. For the linear evaluation (Table \ref{tab:ssv2-linear}) within the same dataset, VideoMoCo stands out by leading in both Top-1 and Top-5 accuracy metrics. Comparing the outcomes for SSv2 across both evaluation methods, the linear evaluation showcases a broader range of results. The lowest Top-1 accuracy is noted at 7.21\% with MoCo, and the highest reaches 19.66\% with VideoMoCo, indicating a more significant variance than in the full fine-tuning approach, where Top-1 accuracy spans from a minimum of 45.26\% with AVID-CMA to a maximum of 57.43\% with TCLR.

In the EK-100 dataset, under the full fine-tuning scenario (see Table \ref{tab:EK-100-fft}), VideoMoCo achieves the best Top-1 accuracy in verb recognition with a score of 44.53\%. However, RSPNet outperforms other models across all evaluated metrics for both noun and action recognition. In the linear evaluation setting of the EK-100 dataset (see Table \ref{tab:ek-linear}), VideoMoCo secures the highest Top-1 and Top-1 accuracy in verb recognition, whereas TCLR stands out by achieving the best Top-1 and Top-1 accuracy for noun and action recognition. Compared to the results on the SSv2 dataset, the EK-100 dataset shows lower performance in action recognition. Particularly in the linear evaluation, only the TCLR model achieves an action recognition accuracy greater than 1\%, while models like MoCo, Pre-Con, and AVID-CMA achieve nearly 0\% accuracy, with specific scores of 0.01\%, 0.01\%, and 0.02\%, respectively. According to the results of both full fine-tuning and linear evaluation, noun and action recognition are significantly more challenging compared to verb recognition. 

Regarding the template-based performance evaluation, there is a clear difference between both datasets. False templates yield the highest performance in both full fine-tuning and linear evaluation modes in the SSv2 dataset (see Figure \ref{fig:ssv2-templatae}). In the full fine-tuning scenario, the TCLR model outperforms others in handling both true and false templates. Conversely, in the linear evaluation scenario, VideoMoCo leads in performance for both template types. Similar result variation as we observed in overall performance in SSv2 can be observed in template-based performance as well, where linear evaluation scores range from a minimum of 10.05\% for false templates and 4.17\% for true templates in MoCo to a maximum of 22.71\% for false templates and 14.03\% for true templates in VideoMoCo. In the context of full fine-tuning, performance spans from a minimum of 44.16\% for false templates and 43.56\% for true templates in AVID-CMA to a maximum of 58.57\% for false templates and 54.80\% for true templates in TCLR models.

In the EK-100 dataset, true templates show superior performance compared to false templates in both full fine-tuning and linear evaluation settings.  VideoMoCo shows the highest verb recognition accuracy (we only consider verbs as templates in the EK-100 dataset) in full fine-tuning setting, achieving 41.37 \% and 56.39\%, respectively, for both false and true templates. In the context of linear evaluation, VideoMoCo leads in recognizing false templates with a 22.66\% accuracy, whereas CTP, MoCo, and Pre-Con exhibit superior performance in identifying true templates, each achieving a 48.41\% accuracy. MoCo, CTP and Pre-Con could not correctly identify any verb in false templates in the linear evaluation setting. 

The observed disparity in the performance of noun and action recognition compared to verb detection within EK-100 dataset can be attributed to the limitations of the R(2+1)D-18 backbone, which struggles with complex actions. Interestingly, excelling in verb recognition does not necessarily guarantee similar success in recognizing nouns or actions. This indicates that these tasks require different skills, and a model's strength in one area doesn't automatically mean it will excel in another. For example, VideoMoCo does well with verbs but falls short in recognizing nouns, highlighting the challenge of adapting methods geared towards verb recognition to the subtle requirements of noun recognition in dynamic video content. On the other hand, RSPNet and TCLR show promise by performing well in both verb and noun recognition tasks. This suggests that these models have a more flexible or effective way of processing and learning from video data, which helps them meet the varied demands of different recognition tasks. Such versatility is key for creating advanced action recognition systems that can accurately identify both the actions being performed and the objects involved in those actions in complex visual settings.

\section{Discussion and Conclusion}

The study of video content using SSL models has emerged as a dynamic area of research, highlighting the complex challenges and distinct opportunities in this field. Among the methods that we explore, TCLR and VideoMoCo emerge as notable performers, showcasing their robust capabilities across various evaluation conditions. VideoMoCo focuses on capturing robust representations that are insensitive to temporal variations, while TCLR focuses on capturing the temporal variations within video instances. 
Both approaches employ discrimination-based learning objectives and focus on learning high-level cues, which likely contributes to their high performance.

However, finding the best-performing model remains complex, with no single model consistently outperforming across all datasets and conditions. This nuanced landscape of results underscores the inherent complexity and diversity of video analysis tasks, revealing the multifaceted nature of visual understanding and interpretation for contact identification.

A focal point of this exploration is the discernible variance in model performance between the EK-100 and SSv2 datasets. The EK-100 dataset, characterized by its ego-centric video perspectives, presents a formidable challenge that starkly contrasts the nature of the SSv2 dataset. This difference is primarily attributed to the ego-centric composition of EK-100 videos, which diverges significantly from the more generalized content found in the Kinetics-400 dataset. The ego-centric viewpoint captures a first-person perspective, often encapsulating complex, nuanced interactions with the environment that are inherently difficult to model. This complexity is increased by the need to accurately detect both verbs and nouns within these interactions, a task that has proven to be particularly challenging within the EK-100 dataset.

The challenge of detecting human contact between objects in the SSv2 dataset shows a major weakness in current SSL models. Although these models do well in identifying contact in the EK-100 dataset, they struggle with the unique challenges of SSv2, highlighting a shortfall in how they learn. Particularly, when using true templates—specific tests designed to see if the models can recognize physical contacts—the difference in performance is even more obvious. Generally, these models perform poorly in detecting contacts across different datasets, but they perform slightly better in the EK-100 dataset, which has a more complex, first-person perspective. This points to a complex issue: while the models have trouble applying what they've learned about contact detection to different datasets, they show a small improvement in the specific environment of EK-100 compared to the different scenarios in the SSv2 dataset. 

Expanding the range of datasets used in this field, the SAYCam dataset \cite{sullivan2021saycam} stands out as an exciting opportunity for future research. This dataset is unique because it is shown from a child's point of view, capturing a variety of daily interactions and visual scenes. Although SAYCam is rich with diverse and unstructured content, it can be quite complex to analyze. This complexity poses challenges but also offers opportunities for SSL models. Using SAYCam could offer valuable insights into how these models process and understand visuals from a unique and very human perspective. This exploration fits well with the broader objective of improving how these models handle and make sense of varied, real-world visuals.

Among the broader challenges of video analysis, the task of accurately identifying contact interactions stands out as a critical area for advancement. The current discussion illuminates a potential pathway forward: integrating hand detection mechanisms to focus on contact-centric interactions. This approach proposes a targeted refinement of the models' capabilities, focusing on the specific, pivotal moments of physical contact within video sequences. By extracting and emphasizing these moments, models can develop a more refined understanding of interactions, potentially overcoming some of the limitations observed in datasets like EK-100. This strategy underscores a pivotal shift towards more specialized, context-aware models that prioritize the detection of meaningful, interaction-centric visual cues.

One of the major drawbacks of our research is that we only use CNN-based models. It would be worthwhile to investigate other architectures, especially ViT-based models in this aspect. As this field continues to evolve, the insights gain from these explorations will undoubtedly contribute to the development of more advanced, capable models. These models will be better equipped to navigate the complexities of real-world visual environments, marking significant progress in the quest for video interpretation and understanding.

\section*{Acknowledgments}

This research was supported by the joint grant P007 from Mohamed Bin Zayed University of Artificial Intelligence  and the Weizmann Institute of Science. The authors would like to express their sincere gratitude for this generous support, which made the study possible.

\bibliographystyle{splncs04}
\bibliography{mybibliography}

\end{document}